\documentclass{article}

\usepackage[preprint]{neurips_2025}

\usepackage[utf8]{inputenc} 
\usepackage[T1]{fontenc}    
\usepackage{hyperref}       
\usepackage{url}            
\usepackage{booktabs}       
\usepackage{amsmath}
\usepackage{amsfonts}       
\usepackage{nicefrac}       
\usepackage{microtype}      
\usepackage{xcolor}         
\usepackage{amsmath,amssymb}
\usepackage{graphicx}
\usepackage{subcaption}
\usepackage{multirow}
\usepackage{wrapfig}
\usepackage{graphicx}
\usepackage{pifont}
\usepackage{colortbl} 
\usepackage{tabularx}
\usepackage{array}
\usepackage{tikz}
\usetikzlibrary{calc}

\title{SkipVAR: Accelerating Visual Autoregressive Modeling via Adaptive Frequency-Aware Skipping }

\author{
Jiajun Li$^{1,5}$\thanks{Equal contribution. $^\dag$Corresponding authors.} \quad 
Yue Ma$^{2*}$ \quad 
Xinyu Zhang$^{1}$ \quad 
Qingyan Wei$^{3}$ \\ 
Songhua Liu$^{4,5\dagger}$ \quad 
Linfeng Zhang$^{5\dagger}$ \\[0.5em]
\hspace{-2em}
$^1$University of Electronic Science and Technology of China \\ 
$^2$The Hong Kong University of Science and Technology \\
$^3$Central South University \ \ \
$^4$National University of Singapore \ \ \
$^5$Shanghai Jiaotong University \\\vspace{-2em}
}

\begin{document}
\maketitle
\begin{figure}[h]
  \centering
  \includegraphics[width=1.0\linewidth]{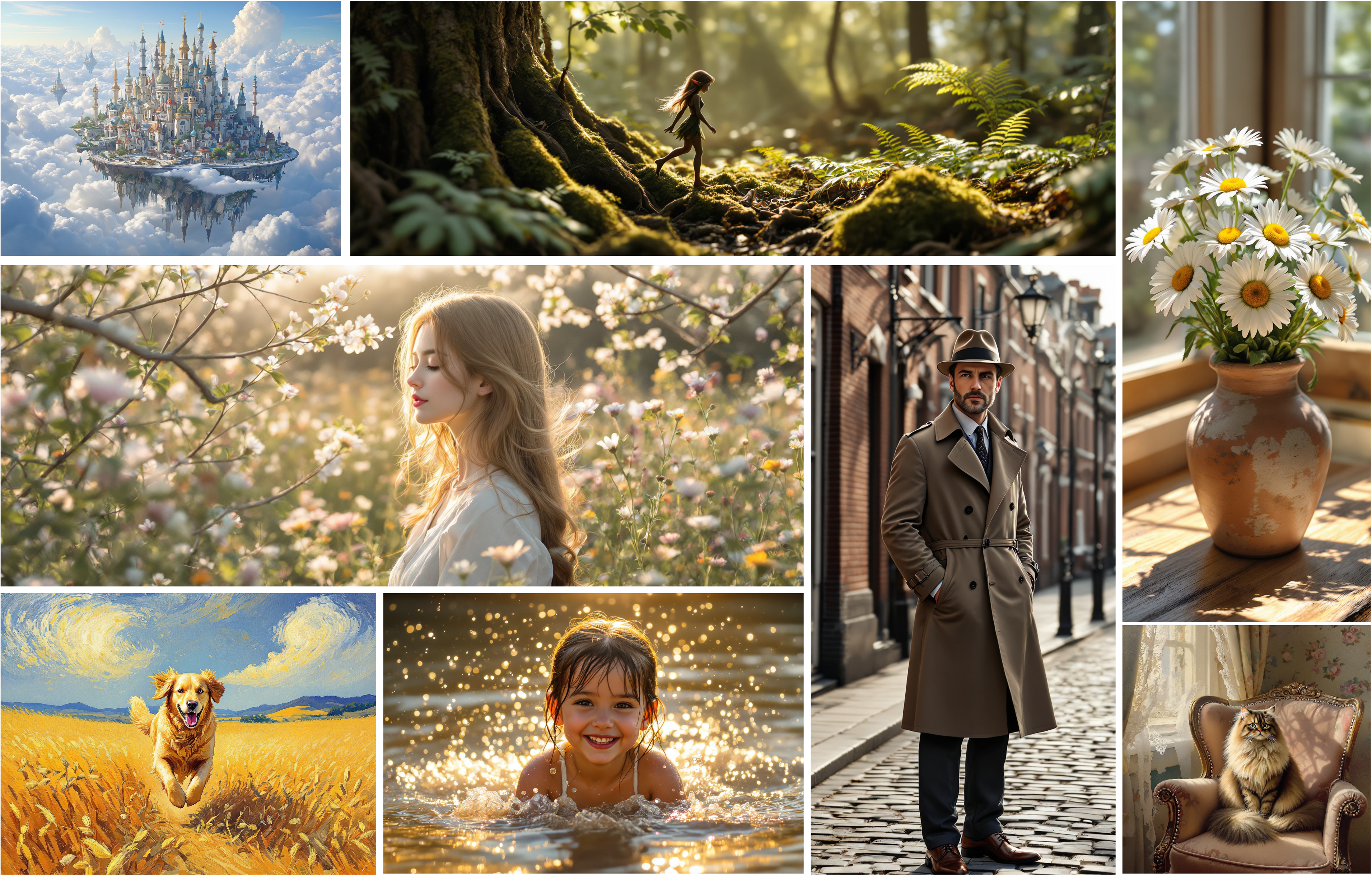}
  \caption{\textbf{Qualitative results of SkipVAR.} We show our SkipVAR based-on Infinity-8B~\cite{han2024infinity}, demonstrating optimal acceleration (average $1.57\times$ speedup) on samples rich in high-frequency content while preserving visual quality.}
  \label{fig_1}
\end{figure}

\begin{abstract}

Recent studies on Visual Autoregressive (VAR) models have highlighted that high-frequency components, or later steps, in the generation process contribute disproportionately to inference latency. 
However, the underlying computational redundancy involved in these steps has yet to be thoroughly investigated. 
In this paper, we conduct an in-depth analysis of the VAR inference process and identify two primary sources of inefficiency: \textit{\textbf{step redundancy}} and \textit{\textbf{unconditional branch redundancy}}. 
To address step redundancy, we propose an automatic step-skipping strategy that selectively omits unnecessary generation steps to improve efficiency.
For unconditional branch redundancy, we observe that the information gap between the conditional and unconditional branches is minimal.
Leveraging this insight, we introduce unconditional branch replacement, a technique that bypasses the unconditional branch to reduce computational cost. 
Notably, we observe that the effectiveness of acceleration strategies varies significantly across different samples. 
Motivated by this, we propose \textbf{SkipVAR}, a sample-adaptive framework that leverages frequency information to dynamically select the most suitable acceleration strategy for each instance. 
To evaluate the role of high-frequency information, we introduce high-variation benchmark datasets that test model sensitivity to fine details. Extensive experiments show SkipVAR achieves over 0.88 average SSIM with up to 1.81x overall acceleration and 2.62x speedup on the GenEval benchmark, maintaining model quality. These results confirm the effectiveness of frequency-aware, training-free adaptive acceleration for scalable autoregressive image generation. 
Our code is available at \url{https://github.com/fakerone-li/SkipVAR} and has been publicly released.
\end{abstract}

\section{Introduction}

Visual Autoregressive (VAR) modeling introduces a novel paradigm in image generation by redefining autoregressive learning as a coarse-to-fine “next-scale prediction” process~\cite{tian2024var}, diverging from the traditional raster-scan “next-token prediction” approach~\cite{li2024mar,liu2024lumina,sun2024llamagen,wang2024emu3,xie2024showo}. This methodology enables autoregressive transformers to efficiently learn visual distributions and generalize effectively. 
\begin{wrapfigure}{l}{0.48\linewidth}
  \centering
  \includegraphics[width=0.95\linewidth]{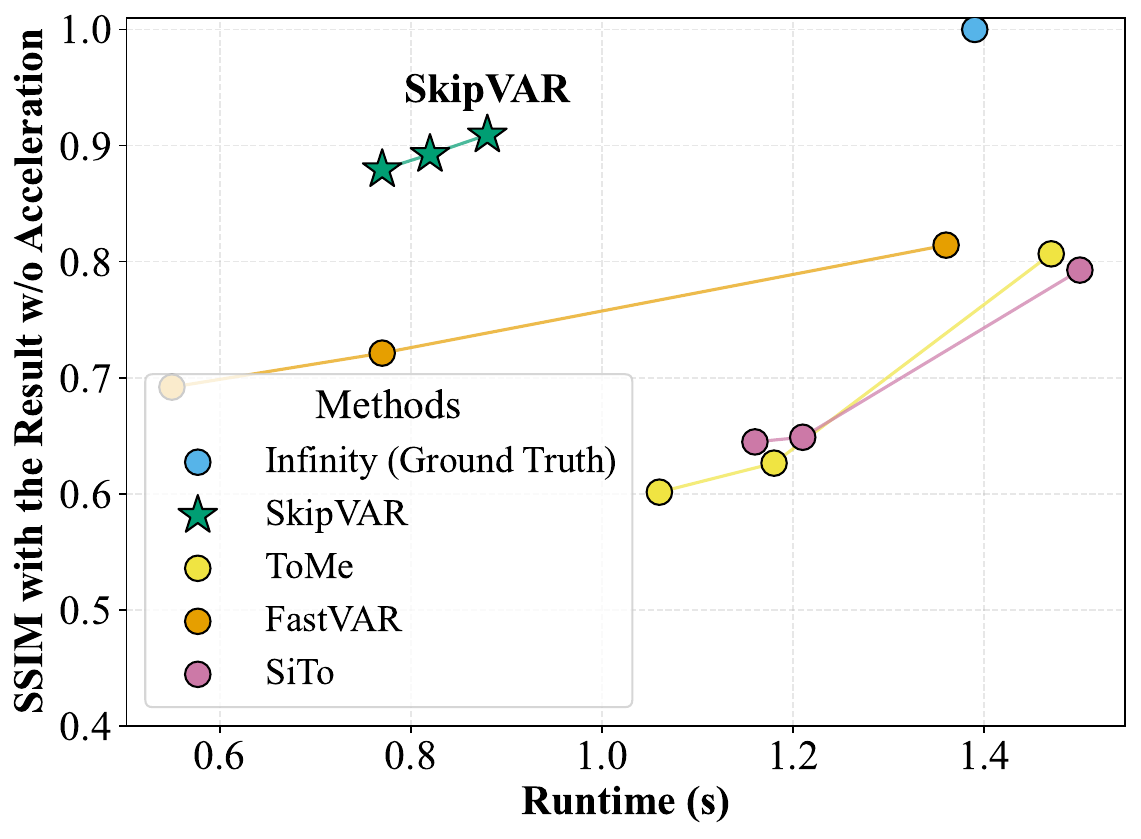}
  \caption{\small \textbf{SSIM v.s. runtime for different acceleration methods.} \textbf{SkipVAR} achieves the best trade-off, maintaining high consistency with faster inference.}
  \label{fig_2}
  \vspace{-1em}
\end{wrapfigure}
Based on the VAR framework, \textbf{Infinity}~\cite{han2024infinity} extends VAR by introducing an infinite‑vocabulary tokenizer and bitwise self‑correction, enabling high‑resolution, photorealistic image generation from text prompts.

Recent advances in accelerating visual autoregressive (VAR) models~\cite{Guo2025FastVAR,chen2024collaborative,Xie2024LiteVAR} have revealed that a significant portion of computational overhead arises from the later stages of generation. These findings are consistent with recent analyses indicating that high-frequency components in the generation process contribute disproportionately to inference latency. Although these components often have limited perceptual significance, they require dense computations, thereby becoming a primary bottleneck in inference efficiency. Consequently, identifying and mitigating the computational redundancy in late-stage generation has emerged as a critical direction for enabling scalable and efficient VAR inference.

\begin{wrapfigure}{l}{0.48\linewidth}
  \vspace{-1em}
  \centering
  \includegraphics[width=\linewidth]{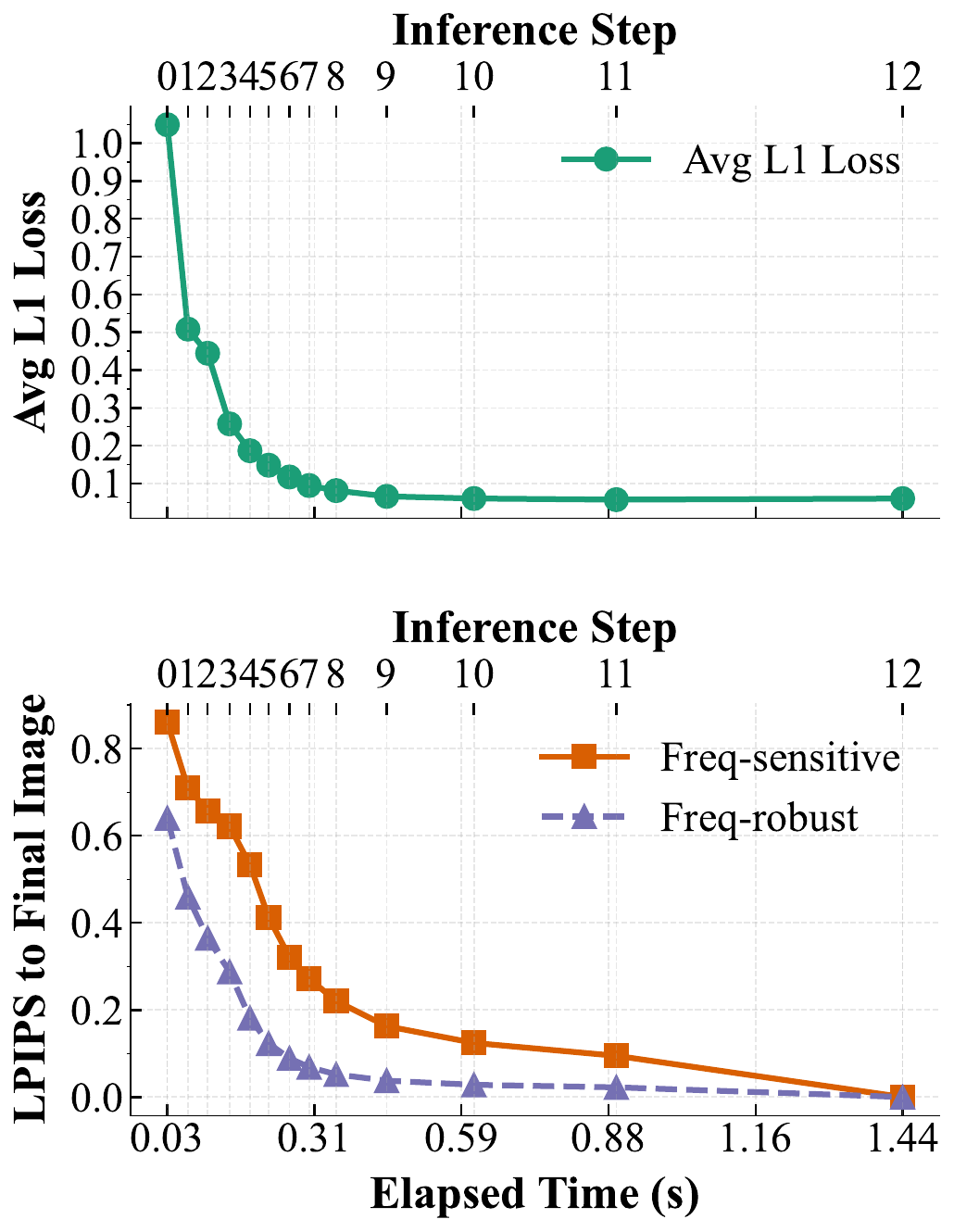}
  
  \caption{\small \textbf{Top:} L1 loss between conditional and unconditional branches drops over inference steps. \textbf{Bottom:} LPIPS to the final image also drops, with frequency-robust samples consistently lower than frequency-sensitive ones.}
  \label{fig:sub-b}
  \vspace{-1em}
\end{wrapfigure}

Based on these observations, a comprehensive understanding of computational inefficiencies in the VAR inference pipeline is still lacking. We conduct an in-depth analysis of the generation dynamics and identify two key forms of redundancy that degrade inference efficiency. The first, \textit{\textbf{step redundancy}}, occurs when later autoregressive steps contribute minimal improvements to the output despite high computational cost, as shown in Figure~\ref{fig:sub-b}. The second, \textit{\textbf{unconditional branch redundancy}}, is found in models with both conditional and unconditional decoding branches, where the information disparity between branches is often negligible, making the unconditional branch largely redundant in later stages. Our acceleration strategy replaces the unconditional branch with conditional-only generation in these stages, significantly reducing computational overhead while maintaining output coherence. Compared to token-level methods~\cite{bolya2022tome,bolya2023tomesd,zhang2024token,zou2024toca}, our approach offers greater stability and consistency.

Notably, we observe that the effectiveness of acceleration strategies varies significantly across different samples. This variability stems from the fact that the importance of high-frequency information differs substantially from one instance to another. Certain images are highly sensitive to fine-grained structural details and thus demand preservation of high-frequency components for faithful reconstruction. Conversely, other samples can tolerate a reduction in high-frequency computation without noticeable quality degradation. Despite this intrinsic heterogeneity, existing VAR acceleration methods predominantly rely on uniform, one-size-fits-all strategies that apply the same acceleration scheme across all samples and throughout the entire inference process. Such approaches overlook the unique frequency characteristics of individual samples, often resulting in suboptimal trade-offs between inference speed and output fidelity. Therefore, there is a clear need for adaptive acceleration frameworks that dynamically tailor the strategy to each sample’s specific frequency requirements.

Motivated by these observations, we propose \textbf{SkipVAR}, a sample-adaptive, decision-driven acceleration framework that dynamically selects the most suitable strategy for each instance based on its frequency characteristics. Unlike prior approaches that apply fixed acceleration schemes, SkipVAR leverages frequency information to tailor inference behavior at the sample level. Specifically, our method employs a lightweight decision model to assess each image’s sensitivity to high-frequency details, and accordingly selects the optimal acceleration policy from a palette of VAR-compatible methods. By fully exploiting the inherent multi-scale structure of VAR models~\cite{tian2024var} and adapting to individual content requirements, SkipVAR ensures that samples requiring high-frequency precision remain unaffected by speedup, while more tolerant samples benefit from substantial computational savings. This dynamic adaptation leads to a significantly improved balance between efficiency and generation quality across diverse visual inputs.



To quantify the role of high-frequency information in autoregressive image generation, we design high-variation benchmark datasets that highlight models' sensitivity to fine-grained details. Given the unreliability of existing subjective metrics in capturing high-frequency discrepancies, we employ objective similarity metrics to assess the perceptual fidelity between original and accelerated outputs. We compute similarity both over the entire image and specifically on high-frequency components, emphasizing the impact of fine details in evaluation. Additionally, we construct two datasets: a \textit{frequency-sensitive} set, where samples are highly affected by high-frequency degradation, and a \textit{frequency-robust} set, where quality remains stable with reduced high-frequency computation. These benchmarks facilitate a more interpretable evaluation of acceleration strategies and underscore the need for sample-adaptive methods like SkipVAR.

Through our SkipVAR method, we achieve superior performance in Figure~\ref{fig_2}, where the curve produced by our approach consistently outperforms a series of token-based methods. Even at a $1.81 \times$ acceleration ratio, we maintain SSIM $>$ 0.88.  Moreover, our approach generalizes seamlessly to the Infinity-8B model, delivering even higher acceleration factors. In Figure~\ref{fig_1}, we present high-quality visual results generated by SkipVAR based on the infinity-8B model, showing the powerful and faster generative ability, including various styles and human faces in various image resolutions. 
Overall, our contributions can be summarized as follows:

\begin{itemize}
    
   \item We introduce two efficient acceleration strategies, step-skipping and unconditional branch replacement, and a high-frequency-focused score to emphasize fine-detail preservation.
   \item We propose SkipVAR,  a frequency-sensitive and sample-adaptive framework that adaptively selects a frequency-aware acceleration strategy for each instance.
    \item To validate the effectiveness of SkipVAR,  we perform extensive experiments and user studies to evaluate our approach, which
     shows our method achieves state-of-the-art performance. 
    
\end{itemize}

\section{Related Work}

\label{sec:related-work}
\paragraph{Visual Autoregressive Models.}
Visual Autoregressive (VAR) models~\cite{tian2024var} depart from traditional token-by-token prediction by adopting a coarse-to-fine, multi-scale next-patch strategy and diffusion transformer~\cite{zhu2024instantswap,chen2024m,ma2025followyourmotion, wang2024cove, zhang2024follow,chen2023attentive,ma2022visual,feng2025dit4edit, zhang2025magiccolor,ma2025follow,yan2025eedit,wan2024unipaint,wang2024taming}: at the earliest scales they capture global layout and structure, and at finer scales they synthesize high-frequency textures and detail. A prominent example, Infinity~\cite{han2024infinity}, extends this paradigm through three key innovations: an infinite-vocabulary tokenizer and classifier enabling bitwise token prediction, a bitwise self-correction mechanism to refine detail accuracy, and a theoretically unbounded scaling of both tokenizer and transformer modules. These advances yield state-of-the-art performance—achieving GenEval and ImageReward (0.96 vs.\ 0.87) scores that surpass SD3-Medium and SDXL—while generating 1024×1024 images in 1.4 s. Nonetheless, the exponential growth of token counts at fine scales leads to substantial computational and memory overhead, and the hierarchical discretization process can accumulate approximation errors that degrade perceptual fidelity.

\paragraph{Acceleration Techniques in Visual Generation.}
Acceleration in visual generation has been explored extensively in diffusion models through distillation~\cite{meng2023distillation,ma2024followyouremoji, ma2024followyourclick, salimans2022progressive}, quantization~\cite{li2023qdiffusion,shang2023ptq4dm}, pruning~\cite{bolya2022tome,bolya2023tomesd,liu2025avatarartist, ma2023followyourpose, xiong2025enhancing, zhu2025multibooth,fang2023structural,wang2024attnprune,zhang2024token,zou2024toca}, and feature caching~\cite{Ma2024DeepCache,Ma2024LearningToCache,liu2024fastericml,li2023fasterdm}. However, these techniques do not directly translate to the hierarchical, patch-based generation of VAR models. FastVAR~\cite{Guo2025FastVAR} addresses this gap by applying post-training token pruning with cached token reuse to accelerate inference without sacrificing quality; yet it employs a fixed pruning ratio across all images and scales, ignoring per-sample variations in high-frequency importance and conditional information and thus risking either wasted computation or perceptual degradation. Our SkipVAR framework departs from these uniform accelerators by dynamically selecting, for each image and at each scale, between aggressive step-skipping and conservative unconditional branch replacement based on handcrafted frequency-sensitivity features. This sample-specific decision mechanism preserves structural coherence, avoids pruning artifacts, and maintains SSIM above a user-specified threshold while delivering efficient inference across diverse inputs.

\section{Methodology}
\subsection{Preliminary}
\begin{figure}[t]
  \centering
  \includegraphics[width=\linewidth]{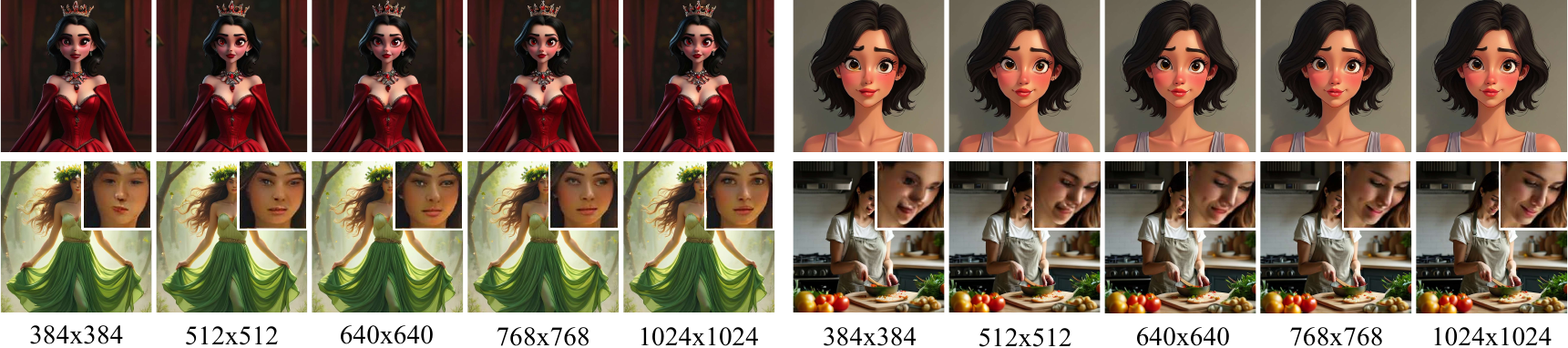}
  \caption{\textbf{Visualization of Post-Scaling Steps for Samples with Varying Sensitivity.}}
  \label{fig:sub-a}
  \vspace{-1em}
\end{figure}
Autoregressive (AR) models factorize the joint probability of a token sequence $\mathbf{x}=(x_1,\dots,x_T)$ as $p(\mathbf{x}) = \prod_{t=1}^T p(x_t \mid x_{<t})$, and are trained to predict each token from its predecessors. Extending AR to images requires discretizing continuous pixels via a VQ-VAE, which encodes an image $\mathrm{im}\in\mathbb{R}^{H\times W\times3}$ to $f=E(\mathrm{im})$, then quantizes it to $q=Q(f)$ with a codebook $Z\in\mathbb{R}^{V\times C}$, and reconstructs the image as $\hat{\mathrm{im}}=D(\mathrm{lookup}(Z,q))$, typically under a combination of reconstruction, perceptual, feature, and adversarial losses. However, the raster-scan autoregressive process breaks spatial locality, incurs $O(n^2)$ decoding steps, and results in $O(n^6)$ operations for an $n\times n$ image. To address this, Visual Autoregressive Modeling (VAR) instead generates coarse-to-fine token maps $\mathbf{R}=(r_1,\dots,r_K)$, where each $r_k$ is a token grid in $[V]^{h_k\times w_k}$ and the final resolution satisfies $(h_K,w_K)=(h,w)$. The probability is modeled as $p(\mathbf{R}) = \prod_{k=1}^K p(r_k\mid r_{<k})$, enabling block-wise causal attention during training and parallel decoding with KV-caching at inference time. This reduces the number of iterations to $O(\log n)$ and computation to $O(n^4)$. Despite these improvements, the rapid increase of fine-scale tokens at high resolutions remains a major computational bottleneck.

\subsection{Motivation}

Visual Autoregressive (VAR) models~\cite{tian2024var} adopt a coarse-to-fine, multi-scale generation paradigm that improves image quality but increases computational cost at higher scales. This section outlines key motivations for our adaptive approach by examining limitations in current methods.

\paragraph{1. Redundancy in High-Frequency Generation.}  
Early scales capture global structure, while later scales refine high-frequency details~\cite{tian2024var,Guo2025FastVAR}. However, such refinement is costly and yields diminishing perceptual returns~\cite{Guo2025FastVAR}. Figure~\ref{fig:sub-b} shows that although later inference steps dominate runtime, LPIPS improvements plateau and image changes become visually minor. Both conditional and unconditional losses decrease steadily, but separately computing these branches imposes significant overhead without commensurate quality gains.

\paragraph{2. Variability in High-Frequency Sensitivity.}  
Images differ in their dependence on high-frequency detail. For instance, an anime-style headshot exhibits minimal changes across steps, while a realistic portrait requires ongoing refinement (Figure~\ref{fig:sub-a}). Existing VAR acceleration methods~\cite{Guo2025FastVAR,chen2024collaborative,Xie2024LiteVAR} apply uniform strategies, risking wasted computation or quality loss. Figure~\ref{fig:sub-b} shows that frequency-robust samples have consistently lower LPIPS than frequency-sensitive ones, highlighting the need for sample-specific acceleration.

\paragraph{3. Limitations of Existing Benchmarks.}  
Common metrics like ImageReward~\cite{xu2023imagereward} and CLIP Score~\cite{hessel2021clipscore} emphasize overall plausibility but neglect fine-detail fidelity crucial in later VAR stages. Consequently, blurred or less detailed outputs may still score highly, masking acceleration effects. We advocate comparing accelerated outputs to full-step references using objective metrics SSIM and LPIPS~\cite{zhang2018unreasonable}, supplemented by high-frequency variants (\texttt{SSIM-HF}, \texttt{LPIPS-HF}) to better evaluate fine-detail preservation.

In sum, redundancy in high-frequency generation, variability in sample sensitivity, and benchmark limitations motivate an adaptive, image-specific acceleration strategy. Our proposed SkipVAR addresses these challenges by dynamically tailoring generation per image, enhancing computational efficiency without compromising perceptual quality.
\begin{figure}[t]
  \centering
  \includegraphics[width=1.0\linewidth]{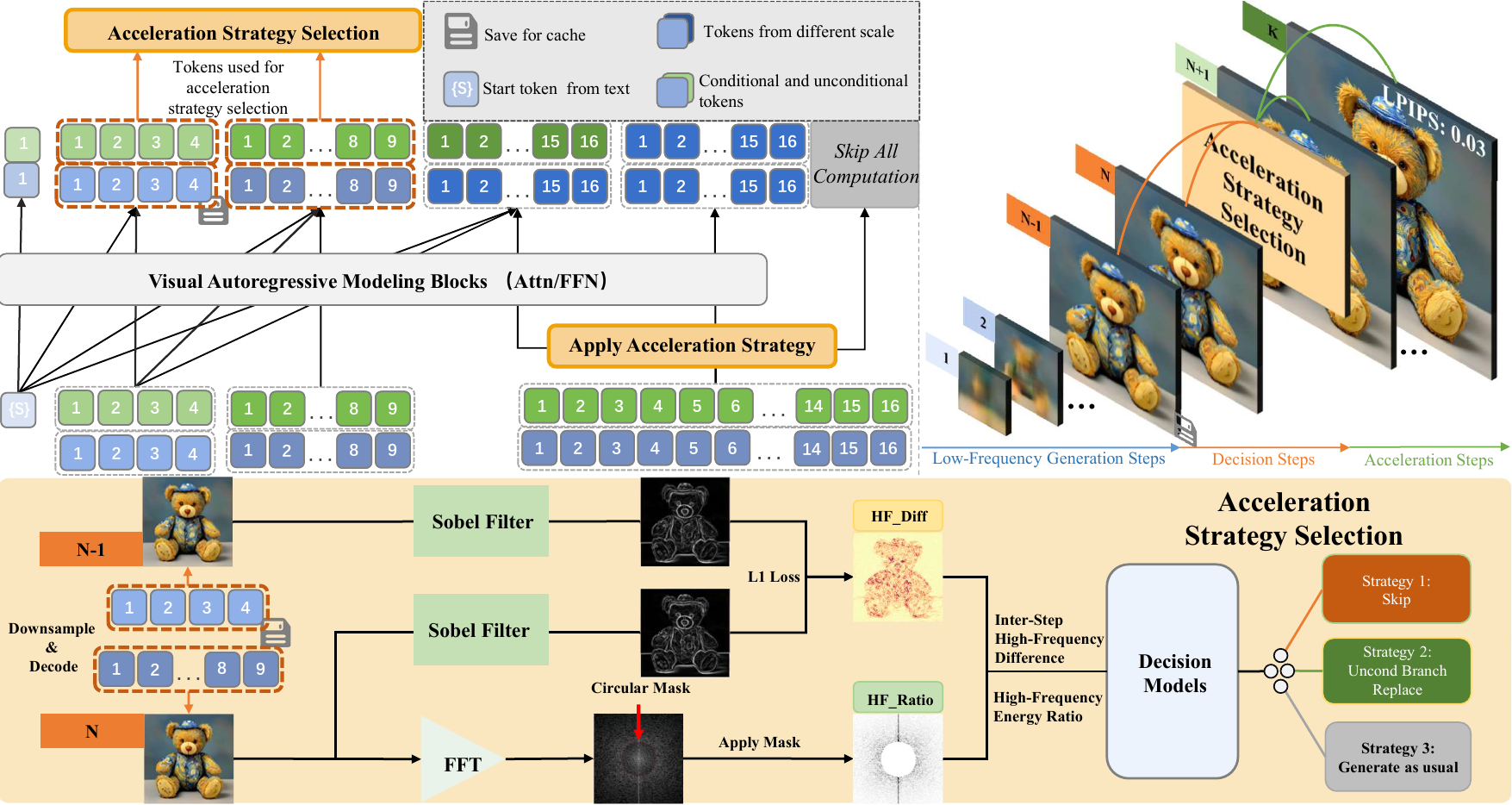}
  \caption{\textbf{Top Left:} Overall framework. At the Decision Step, an acceleration strategy is selected and applied to subsequent steps. For clarity in showcasing the acceleration strategies, we omit the propagation of other methods during the acceleration steps.  
\textbf{Top Right:} Visualization of SkipVAR's pipeline. We divide the generation process into \textit{Low-Frequency Generation Steps}, \textit{Decision Steps}, and \textit{Acceleration Steps} to emphasize the roles of each stage. At Decision Step, we cache the results from the previous step, then select an appropriate acceleration strategy to apply during the subsequent Acceleration Steps.  
\textbf{Bottom:} Detailed framework for acceleration strategy selection. After downsampling and decoding the results into image space, we compute $\mathrm{HF\_Diff}$ by applying the Sobel filter and measuring the L1 loss, capturing the difference in high-frequency energy across steps. Simultaneously, we compute $\mathrm{HF\_Ratio}$ by applying a circular mask in the frequency domain via FFT to measure the proportion of high-frequency content. These indicators are then fed into a decision model to determine the optimal acceleration strategy.}
  \label{fig:SkipVAR}
  \vspace{-1em}
\end{figure}
\subsection{Method}

We propose \textbf{SkipVAR}, a post-training adaptive acceleration framework designed for Visual Autoregressive (VAR) models. SkipVAR exploits the inherent redundancy in multi-scale generation by dynamically selecting acceleration strategies on a per-sample basis, guided by the significance of high-frequency components. Our method aims to substantially reduce inference cost without compromising perceptual quality.

\paragraph{Overview of SkipVAR.} 
Given a multi-scale VAR model that generates an image over $K$ refinement steps, we designate a specific step $N$ (typically early, e.g., $N=9$) as the \textbf{decision step}. After step $N$, acceleration strategies may be selectively applied to the remaining steps $\{N+1, N+2, \dots, K\}$. To inform decision-making, we utilize decoded images from step $N$ and the cached intermediate result from step $N{-}1$. These decoded outputs are downsampled and analyzed in the image space to estimate the importance of high-frequency information. The extracted features are then fed into a lightweight decision model $\mathcal{D}$, which outputs the optimal acceleration policy for the remaining steps on a per-sample basis.

\paragraph{High-Frequency Importance Estimation.} 
To characterize a sample’s sensitivity to high-frequency information, we compute two complementary indicators:

\subparagraph{High Frequency Difference (HF\_Diff).}
To quantify local variations in high-frequency structures during the generation process, we apply a $3 \times 3$ Sobel operator $\mathcal{S}$ to the grayscale images decoded at two consecutive steps, $N$ and $N{-}1$. The high-frequency difference is then defined as the $\ell_1$ distance between their corresponding Sobel responses:

\begin{equation}
\mathrm{HF\_Diff} = \left\| \mathcal{S}(I_N) - \mathcal{S}(I_{N-1}) \right\|_1,
\label{eq:hf_diff}
\end{equation}

where $I_N$ and $I_{N-1}$ denote the decoded grayscale images at steps $N$ and $N{-}1$, respectively. A smaller value of $\mathrm{HF\_Diff}$ indicates that the high-frequency content has stabilized between the two steps, implying diminishing returns from further refinement. Therefore, when $\mathrm{HF\_Diff}$ falls below a threshold, the sample is considered less sensitive to high-frequency variations, making it a strong candidate for early termination or aggressive acceleration.

\subparagraph{High Frequency Ratio (HF\_Ratio).}
To complement the local difference measurement, we further assess the global frequency composition of the decoded image using the Fourier transform. Specifically, we define the high-frequency ratio as:

\begin{equation}
\mathrm{HF\_Ratio} = \frac{\sum_{(u,v)\in\mathcal{H}} \left|\mathcal{F}(I_N)\right|_{uv}}{\sum_{(u,v)} \left|\mathcal{F}(I_N)\right|_{uv} + \epsilon},
\label{eq:hf_ratio}
\end{equation}

where $\mathcal{F}(I_N)$ denotes the magnitude of the shifted Fourier transform of the grayscale image $I_N$, and $\mathcal{H}$ represents the set of high-frequency components located beyond a radius $\rho$ (typically $\rho = 0.25$) from the center of the frequency spectrum. The small constant $\epsilon$ is added to avoid division by zero. A lower $\mathrm{HF\_Ratio}$ implies that the image contains inherently less high-frequency information, making it more robust to early truncation or simplified refinement. Therefore, such images are strong candidates for aggressive acceleration.

\paragraph{Lightweight Decision Models.}
To achieve low-latency inference decisions, we employ simple, efficient classifiers for the decision model $\mathcal{D}$. Specifically, we consider:
1. \textbf{Decision Tree (DT):} A tree-based model that partitions the feature space with learned thresholds, offering fast and interpretable decisions.
2. \textbf{Random Forest (RF):} An ensemble of decision trees that aggregates predictions across multiple randomized subsets to improve robustness and generalization.
3. \textbf{Logistic Regression (LR):} A linear model that estimates the probability of applying acceleration strategies via a sigmoid activation over the feature vector.

\paragraph{Dataset Construction.}
We construct the training dataset from the \textbf{People} class, which inherently includes both high-frequency–insensitive samples (e.g., stylized headshots) and high-frequency–sensitive ones (e.g., photo-realistic full-body images). This heterogeneity ensures coverage of acceleration behaviors ranging from aggressive skipping to conservative generation with no acceleration. For each image, we extract two handcrafted features, $\mathrm{HF\_Diff}$ and $\mathrm{HF\_Ratio}$, capturing local Sobel edge stability and global high-frequency content in the Fourier domain, respectively. Together, these features provide a compact yet informative representation of sample frequency sensitivity.

\paragraph{Labeling and Classifier Training.}
Labels are assigned based on the most aggressive acceleration strategy that maintains SSIM above a target threshold, either by skipping steps or replacing unconditional branch with conditional copying. Feature vectors are standardized, and classifiers are trained using an 80/20 train/validation split. While multiple decision models are evaluated, all subsequent experiments use Logistic Regression, highlighting that the framework’s effectiveness stems from its strategic design rather than model complexity.

\paragraph{Generalization Evaluation.}
To validate generalization, we apply the trained decision model to a distinct image category with different content distribution. As shown in Figure~\ref{fig:selection}, SkipVAR remains effective without retraining, achieving an SSIM near 0.88 using a 0.84 threshold—close to the non-accelerated baseline. This suggests that our frequency-aware decision mechanism captures generalizable patterns beyond the training domain.

\paragraph{Acceleration Strategies.}
SkipVAR incorporates two practical strategies to exploit redundancy at late generation stages:

\begin{itemize}
    \item \textbf{Skip Strategy:} For the Skip Strategy, we target frequency-robust samples by selecting a specific acceleration step beyond which decoding stops early. As illustrated in Figure~\ref{fig:SkipVAR}, we directly decode the output at that step without proceeding to further resolution refinements.
    \item \textbf{Uncond-Branch-Replace Strategy:} For the unconditional branch replacement Strategy, we focus on frequency-sensitive samples. As shown in Figure~\ref{fig:sub-b}, we leverage the convergence between the conditional and unconditional branches in the later stages. By reusing the conditional output in place of the unconditional one, we effectively cut computation in half. Compared to the Skip Strategy, the additional cost of computing the conditional branch ensures that frequency-sensitive samples still receive sufficient high-frequency information.
\end{itemize}

Each strategy is governed by an independent decision model: $\mathcal{D}_{\text{skip}}$ predicts steps where skipping is safe, and $\mathcal{D}_{\text{uncond}}$ selects steps where unconditional branch replacement is appropriate.

\begin{table}[t]
  \caption{\textbf{Quantitative comparison on the DrawBench dataset.}
    Note that \textbf{ToMe} and \textbf{SiTo} use the default ratio of \{0.5, 0.5, 0.5\},
    while \textbf{FastVAR} is reproduced by us using official settings.}
  \label{tab:ssim-study}
  \centering
  \resizebox{0.9\linewidth}{!}{
  \begin{tabular}{lccccc}
    \toprule
    Methods & Speedup & SSIM & SSIM-HF & LPIPS & LPIPS-HF \\
    \midrule

    \rowcolor{cyan!10}
    Infinity                    & $1.00\times$ & -      & -      & -      & -      \\
    
    \rowcolor{green!15}
    +SkipVAR@0.88              & $1.58\times$ & 0.9092 & 0.3429 & 0.0488 & 0.1823 \\
    \rowcolor{green!15}
    +SkipVAR@0.86              & $1.70\times$ & 0.8924 & 0.3100 & 0.0577 & 0.1925 \\
    \rowcolor{green!15}
    +SkipVAR@0.84              & $1.81\times$ & 0.8793 & 0.2881 & 0.0646 & 0.1997 \\
    \rowcolor{green!15}
    +SkipVAR@0.84\_HFDiffOnly  & $1.72\times$ & 0.8747 & 0.2815 & 0.0671 & 0.2019 \\
    
    \rowcolor{orange!15}
    +ToMe@0.05\_LastStepOnly   & $0.95\times$ & 0.8067 & 0.1913 & 0.1043 & 0.2376 \\
    \rowcolor{orange!15}
    +ToMe                      & $1.18\times$ & 0.6264 & 0.1111 & 0.3520 & 0.4133 \\
    
    \rowcolor{yellow!15}
    +SiTo@0.05\_LastStepOnly   & $0.93\times$ & 0.7927 & 0.1816 & 0.1092 & 0.2412 \\
    \rowcolor{yellow!15}
    +SiTo                      & $1.15\times$ & 0.6487 & 0.1168 & 0.2874 & 0.3670 \\
    
    \rowcolor{red!15}
    +FastVAR@0.05\_LastStepOnly& $1.02\times$ & 0.8142 & 0.1964 & 0.0977 & 0.2330 \\
    \rowcolor{red!15}
    +FastVAR                   & $2.53\times$ & 0.6919 & 0.1270 & 0.2042 & 0.3092 \\
    
    \rowcolor{green!15}
    +SkipVAR-hybrid (w/o DM)   & $2.62\times$ & 0.7915 & 0.1829 & 0.1075 & 0.2385 \\
    
    \bottomrule
  \end{tabular}
  \vspace{-3em}
  }
\end{table}
\section{Experiments}

\subsection{Experiments Setup}

\paragraph{Setup and Acceleration Mechanism.}  
We evaluate our proposed acceleration strategy on the open-source Infinity-2B model, a state-of-the-art Bitwise Visual AutoRegressive (VAR) model capable of generating high-resolution images up to 1024×1024 pixels. The acceleration decision in SkipVAR is made at the 9th inference step, based on two handcrafted frequency-aware features: the high-frequency difference (computed using a third-order Sobel operator) and the high-frequency ratio (thresholded with $\rho=0.4$). A decision model is trained on the \textit{people} category of the MJHQ30K dataset~\cite{li2024mjhq}, which consists of 3,000 diverse high-resolution images including both stylized and photo-realistic human figures. This model selects the most aggressive yet fidelity-preserving acceleration strategy—either skipping or unconditional replacement—subject to SSIM thresholds $\{0.88, 0.86, 0.84\}$.

\begin{wraptable}{r}{0.48\linewidth}
  \centering
  \vspace{-1em}
  \scriptsize
  \setlength{\tabcolsep}{2pt}
  \renewcommand{\arraystretch}{0.9}
  \caption{\small \textbf{Evaluation on different datasets.}  
    \textcolor{gray!70}{Gray: DrawBench},  
    \textcolor{orange!80!black}{Orange: Paintings},  
    \textcolor{cyan!80!black}{Cyan: Photo}.}
  \label{tab:imagereward}
  \vspace{-0.5em}
  \begin{tabular}{lccccc}
    \toprule
    Method & Spd. & MACs & IRwd & Clip \\
    \midrule
    \rowcolor{gray!10} Infinity        & 1.00× & 31097 & 0.8618 & 0.2702 \\
    \rowcolor{gray!10} +SkipVAR@0.88  & 1.58× & 18134 & \textcolor{red}{0.8688} & 0.2701 \\
    \rowcolor{gray!10} +SkipVAR@0.86  & 1.70× & 16467 & 0.8570 & 0.2705 \\
    \rowcolor{gray!10} +SkipVAR@0.84  & \textcolor{red}{1.81×} & \textcolor{red}{15237} & 0.8575 & 0.2704 \\
    \rowcolor{orange!10} Infinity     & 1.00× & --    & 1.2767 & 0.2746 \\
    \rowcolor{orange!10} +SkipVAR@0.84& 1.53× & --    & 1.2742 & 0.2745 \\
    \rowcolor{cyan!8} Infinity        & 1.00× & --    & 0.9991 & 0.2618 \\
    \rowcolor{cyan!8} +SkipVAR@0.88   & 1.55× & --    & 0.9876 & 0.2621 \\
    \rowcolor{cyan!8} +SkipVAR@0.84   & 1.75× & --    & 0.9854 & 0.2625 \\
    \bottomrule
  \end{tabular}
  \vspace{-1em}
\end{wraptable}
\paragraph{Evaluation Protocol and Metrics.} 
During evaluation, acceleration is applied only to autoregressive decoding steps (10–12), excluding VAE decoding time for fair comparisons. We report generation quality using ImageReward~\cite{xu2023imagereward}, GenEval~\cite{ghosh2023geneval}, HPSv2~\cite{wu2023human}, GPT Scores~\cite{achiam2023gpt}, SSIM~\cite{wang2004image}, LPIPS~\cite{zhang2018unreasonable}, and FID~\cite{heusel2017gans}. Efficiency is measured by Acceleration Ratio, Inference Time, and FLOPS. Experiments are conducted on a single NVIDIA RTX 4090D GPU with 24GB VRAM, using default settings for consistent comparison.

\subsection{Main Results}

\paragraph{Comparison on subjective scores.}  

\begin{figure}[t]
  \centering
  \includegraphics[width=1.0\linewidth]{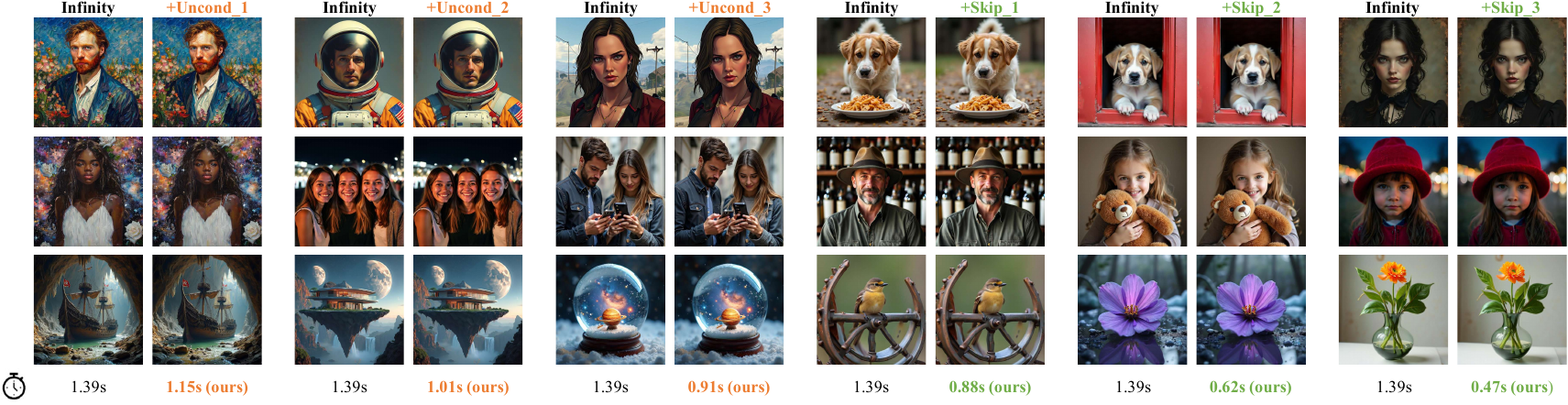}
  \caption{\textbf{Image generations under different acceleration strategies.} Strategies are selected by our decision model. Simpler images use faster strategies with less high-frequency detail; complex ones use slower strategies to preserve more detail. Images are grouped and sorted by strategy speed.}
  \label{fig:samples}
  \vspace{-1.5em}
\end{figure}

\begin{wraptable}{r}{0.48\linewidth}
  \centering
  \vspace{-1em}
  \scriptsize 
  \setlength{\tabcolsep}{2pt} 
  \renewcommand{\arraystretch}{0.9} 
  \caption{\small \textbf{Comparison on multiple benchmarks.} (a) Evaluation on the HPSv2 benchmark. (b) GPT-based evaluation on frequency-sensitive and frequency-robust datasets; \textcolor{blue!80!black}{blue rows} correspond to \textbf{Frequency-sensitive} dataset, \textcolor{green!50!black}{green rows} correspond to \textbf{Frequency-robust} dataset.}
  \vspace{-0.5em}
  \begin{subtable}[t]{\linewidth}
    \centering
    \caption{HPSv2 evaluation.}
    \label{tab:hpsv2-quality}
    \begin{tabular}{p{1.7cm}p{1.1cm}cccc}
      \toprule
      Methods & Speedup & Ani. & Art & Paint & Photo \\
      \midrule
      Infinity       & 1.00$\times$ & 31.70 & 30.45 & 30.40 & 29.43 \\
      +SkipVAR@0.84  & 1.73$\times$ & 31.59 & 30.27 & \textcolor{red}{30.49} & 29.30 \\
      \bottomrule
    \end{tabular}
  \end{subtable}

  \vspace{0.3em}

  \begin{subtable}[t]{\linewidth}
    \centering
    \caption{GPT-based frequency evaluation.}
    \label{tab:skipvar-hf-analysis}
    \begin{tabular}{p{1.7cm}p{0.8cm}p{1.0cm}cccc}
      \toprule
      Methods & Lat. & Spd. & Aes & Align & SSIM \\
      \midrule
      \rowcolor{blue!5}
      Infinity         & 1.39 & --    & 88.36 & 86.69 & --     \\
      \rowcolor{blue!5}
      +SkipVAR@0.84    & 1.09 & 1.28× & 88.11 & \textcolor{red}{87.59} & 0.849 \\
      \rowcolor{green!5}
      Infinity         & 1.39 & --    & 87.18 & 89.75 & --     \\
      \rowcolor{green!5}
      +SkipVAR@0.84    & 0.70 & 1.99× & 87.02 & \textcolor{red}{90.98} & 0.905 \\
      \bottomrule
    \end{tabular}
  \end{subtable}
  \vspace{-2em}
\end{wraptable}
1. \textbf{ImageReward and CLIP Score Evaluation:} We evaluate SkipVAR on DrawBench (200 prompts) and on the Paintings and Photo subsets of HPSV2~\cite{wu2023human} (800 images each) using ImageReward~\cite{xu2023imagereward} and CLIP Score~\cite{hessel2021clipscore}. As shown in Table~\ref{tab:imagereward}, SkipVAR@0.84, SkipVAR@0.86, and SkipVAR@0.88 all achieve near‐identical scores to the original Infinity model across all three datasets, demonstrating that our method maintains high subjective quality even on stylistically diverse image sets.  
2. \textbf{Quality–Speed Tradeoff on HPSV2:} On the full HPSV2 benchmark (Table~\ref{tab:hpsv2-quality}), SkipVAR@0.84 attains a $1.73\times$ speedup for only a 0.08‐point drop in average score (from 30.49 to 30.41), preserving Anime, Concept‐art, Paintings, and Photo quality within 0.2 points. Figure~\ref{fig:samples} visualizes representative outputs ordered by acceleration level, showing that simpler inputs trigger aggressive skipping while complex scenes retain unconditional branch replacement in line with human perceptual expectations.  
3. \textbf{GenEval Benchmark Performance:} On the GenEval benchmark~\cite{ghosh2023geneval} (Table~\ref{tab:geneval}), SkipVAR@0.86 delivers a $1.77\times$ speedup with only a $\sim$1\% drop in overall score (from 0.71 to 0.70). Furthermore, our SkipVAR‐hybrid (w/o DM) configuration—applying skip to steps 11–12 and unconditional branch replacement to steps 9–10—achieves a $2.62\times$ speedup while matching or exceeding baseline two‐objective, position, and color attribute scores, underscoring the robustness of our frequency‐aware acceleration design.
4. \textbf{Adaptation to Frequency Sensitivity:} We categorize HPSV2 images into \emph{frequency‐sensitive} and \emph{frequency‐robust} sets based on SSIM sensitivity at late inference steps. Table~\ref{tab:skipvar-hf-analysis} shows that SkipVAR@0.84 applies the slower unconditional branch replacement strategy for sensitive samples—preserving SSIM at 0.8493—and the faster skip strategy for robust samples—achieving SSIM of 0.9051—yielding up to a $1.99\times$ speedup without perceptible subjective degradation. Additionally, we observe that the GPT score does not decrease due to acceleration and even shows a slight increase in the Align metric, indicating that our approach maintains high semantic consistency while achieving significant speedup.

\newcommand{\diffusion}{\textcolor{blue}{\ding{108}}}       
\newcommand{\autoregressive}{\textcolor{orange}{\ding{110}}} 
\newcommand{\base}{\textcolor{teal}{\ding{72}}}              
\begin{table}[t]
  \caption{\textbf{Comparison of different generation models on the GenEval~\cite{ghosh2023geneval} benchmark.}
  \diffusion\ Diffusion models, \autoregressive\ Autoregressive models, \base\ Infinity model.
  Prompt rewriting was applied to the Infinity model in our experiments.}
  \label{tab:geneval}
  \centering
  \renewcommand{\arraystretch}{1.1}
  \resizebox{\linewidth}{!}{ 
  \begin{tabular}{l c c c c c c}
    \toprule
    Methods & Params & Speedup & Two Obj. & Position & Color Attri. & Overall \\
    \midrule
    \diffusion SDXL                        & 2.6B & -            & 0.74 & 0.15 & 0.23 & 0.55 \\
    \diffusion PixArt-Sigma                & 0.6B & -            & 0.62 & 0.14 & 0.27 & 0.55 \\
    \diffusion SD3-medium                  & 2.0B & -            & 0.74 & 0.34 & 0.36 & 0.62 \\
    \autoregressive LlamaGen               & 0.8B & -            & 0.34 & 0.07 & 0.04 & 0.32 \\
    \autoregressive Show-o                 & 1.3B & -            & 0.80 & 0.31 & 0.50 & 0.68 \\
    \rowcolor{gray!10} \base Infinity                         & 2.0B & -            & 0.85 & 0.39 & 0.55 & 0.71 \\
    \rowcolor{gray!10} \base +SkipVAR@0.88                    & 2.0B & 1.50$\times$ & 0.83 & 0.39 & \textcolor{red}{0.59} & 0.71 \\
    \rowcolor{gray!10} \base +SkipVAR@0.86                    & 2.0B & 1.77$\times$ & 0.84 & 0.36 & \textcolor{red}{0.58} & 0.70 \\
    \rowcolor{gray!10} \base +SkipVAR-hybrid (w/o DM)         & 2.0B & \textcolor{red}{2.62$\times$} & 0.84 & 0.39 & \textcolor{red}{0.60} & \textcolor{red}{0.72} \\
    \bottomrule
  \end{tabular}
  }
  \vspace{-2em}
\end{table}

\paragraph{Comparison on objective metrics.}  
\begin{wraptable}{r}{0.48\linewidth}
  \vspace{-1em} 
  \centering
  \scriptsize
  \caption{\small \textbf{Generation quality on MJHQ30K}}
  \label{tab:mjhq30k}
  \small
  \begin{tabular}{@{}lllll@{}}
    \toprule
    Dataset & Speedup & FID & SSIM & LPIPS \\
    \midrule
    Food    & 1.88$\times$ & 3.56 & 0.8675 & 0.0780 \\
    Fashion & 1.76$\times$ & 3.64 & 0.8713 & 0.0765 \\
    Plants  & 1.60$\times$ & 2.69 & 0.8690 & 0.0649 \\
    Art     & 1.58$\times$ & 3.02 & 0.8663 & 0.0704 \\
    \bottomrule
  \end{tabular}
  \vspace{-1em}  
\end{wraptable}
Because subjective scores often fail to capture high‐frequency detail differences, we evaluate our method using rigorous objective metrics as follows (further discussion is provided in the Appendix):    
1. \textbf{SSIM and LPIPS:} We assess acceleration effects with SSIM and LPIPS under a fixed random seed for direct comparability to the Infinity baseline. On DrawBench (Table~\ref{tab:ssim-study}), SkipVAR maintains SSIM above each model’s training threshold and outperforms token‐based approaches such as ToMe~\cite{bolya2023tomesd}, SiTo~\cite{zhang2024token}, and FastVAR~\cite{Guo2025FastVAR}. For instance, ToMe@0.05\_LastStepOnly—applying 5\% token merging at the final step—yields SSIM < 0.81, whereas SkipVAR’s holistic acceleration better preserves global fidelity. 2. \textbf{SSIM‐HF Analysis:} To capture fine‐detail fidelity, we introduce \texttt{SSIM-HF}, computed on high‐frequency regions only. Also on DrawBench (Table~\ref{tab:ssim-study}), ToMe@0.05\_LastStepOnly shows overall SSIM just 8\% lower than SkipVAR@0.84, but \texttt{SSIM-HF} degrades by 34\%, demonstrating the severe impact of suboptimal token selection on fine textures. 3. \textbf{MJHQ30K FID Performance:} On the MJHQ30K benchmark (Table~\ref{tab:mjhq30k}), SkipVAR@0.84 achieves a $1.88\times$ speedup with FID = 3.5627. Even on the “Art” subset—where high‐frequency detail is paramount—it delivers a $1.58\times$ acceleration without significant quality loss.

\subsection{Empirical Studies}

\begin{figure}[t]
  \centering

  \begin{subfigure}[b]{0.48\linewidth}
    \centering
    \includegraphics[width=\linewidth]{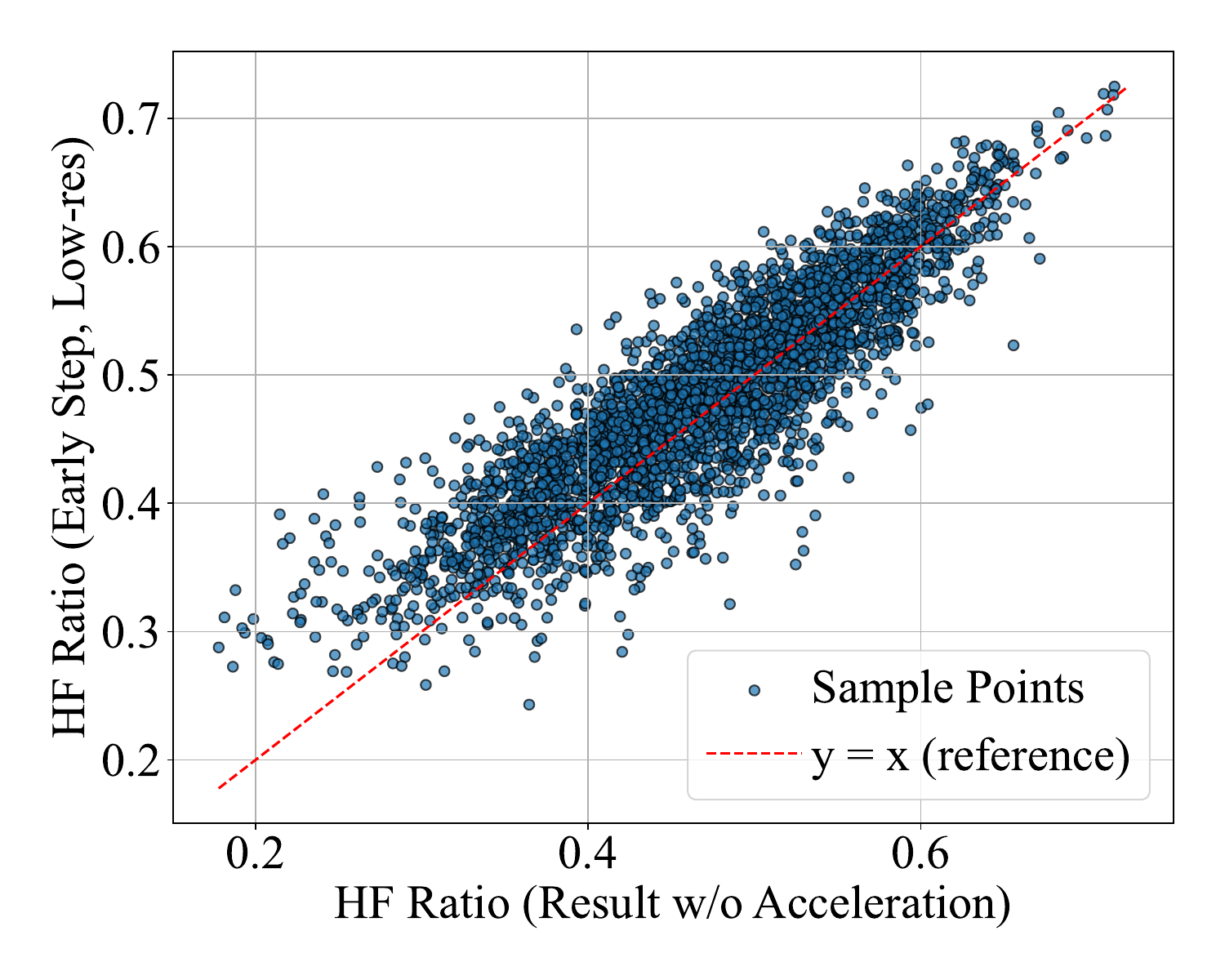}
    \vspace{-2em}
    \caption{$\mathrm{HF\_Ratio}$ at step 9 vs.\ final output}
    \label{fig:hf-ratio}
  \end{subfigure}
  \hfill
  \begin{subfigure}[b]{0.48\linewidth}
    \centering
    \includegraphics[width=\linewidth]{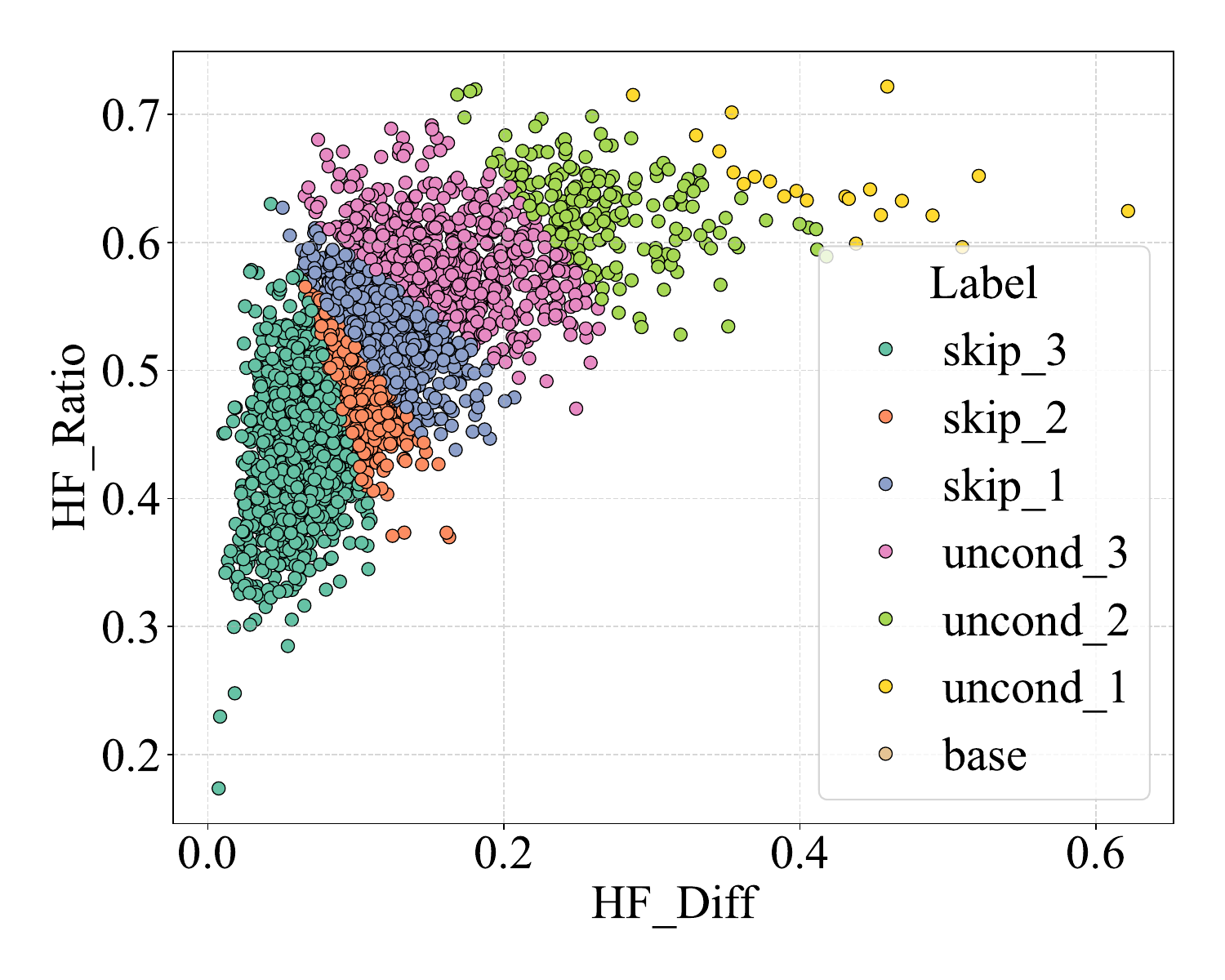}
    \vspace{-2em}
    \caption{Decision regions on Animals class}
    \label{fig:selection}
  \end{subfigure}

  \caption{\textbf{(a)} Scatter of $\mathrm{HF\_Ratio}$ on the 9th-step downsampled image versus the final high-res output—points lie along $y\!=\!x$, showing reliable low-res Fourier estimates.  
\textbf{(b)} Logistic regression boundary in $\mathrm{HF\_Diff}$–$\mathrm{HF\_Ratio}$ space for the Animals class.  
\textbf{Legend:} \texttt{skip\_3} = skip final 3 steps; \texttt{uncond\_2} = uncond branch replace for last 2 steps.}

  \label{fig:dual-fig}
\end{figure}

\paragraph{Feature Design and Objective Analysis.}  
Robust decision-making requires features that correlate with fine-detail sensitivity yet resist luminance and contrast variations. Sole reliance on Sobel-based $\mathrm{HF\_Diff}$, which captures dynamic stability of high-frequency content between refinement steps, can misclassify dark but smooth images as high-frequency. To address this, we introduce $\mathrm{HF\_Ratio}$, which measures the intrinsic frequency distribution of the current sample, reflecting its global reliance on fine-scale details. Combining $\mathrm{HF\_Diff}$ and $\mathrm{HF\_Ratio}$ yields a two-dimensional feature vector that captures both local edge stability across steps and global texture richness. As Table~\ref{tab:ssim-study} shows, SkipVAR@0.84 outperforms the $\mathrm{HF\_Diff}$-only variant (SkipVAR@0.84\_HFDiffOnly) by avoiding excessive conservatism in acceleration decisions. Figure~\ref{fig:hf-ratio} further confirms that the intermediate $\mathrm{HF\_Ratio}$ reliably estimates final high-frequency content, justifying its use at the decision step.

\paragraph{High-Frequency Sensitivity Prediction.}  
We train a logistic regression (LR) on these features to classify samples into “skip” or “uncond branch replace” regimes. As Figure~\ref{fig:selection} illustrates, LR cleanly divides the feature space, yielding SSIM 0.8628 on the Animals class. Decision tree and random forest classifiers achieve comparable accuracy with sub-0.03 ms inference latency, indicating that our handcrafted features carry the predictive burden. Trained on 80\% of the 3K-sample People dataset, all models generalize across MJHQ30K classes and transfer without loss to an 8B-parameter variant, where SkipVAR@0.84 attains a $2.06\times$ speedup and SSIM 0.86 on DrawBench (additional 8B results in Appendix).We note, however, that training on only 3K samples may lead to mild overfitting in ensemble methods such as RF.

\paragraph{Acceleration Program.}
Profiling reveals that the last few decoding steps (10–12) dominate runtime, despite most token distributions having stabilized. We thus fix the \emph{decision point} at step 9 ($N=9$), and only accelerate the final three steps—striking a balance between computational efficiency and structural completeness. This design avoids redundant high-resolution decoding while preserving image layout. Steps 10–12 account for 69\% of total inference time, and LPIPS improvements taper off beyond step 9, even for frequency-sensitive cases (see Figure~\ref{fig:sub-b}). This suggests a natural turning point where acceleration yields diminishing perceptual returns. As shown in Table~\ref{tab:mjhq30k}, SkipVAR achieves up to $1.88\times$ speedup on datasets like \textit{Food} (3K samples), while maintaining SSIM above 0.86. A detailed breakdown of our acceleration strategies is provided in the Appendix.

\section{Conclusion}

In this paper, we presented SkipVAR, a novel, training-free acceleration framework for Visual Autoregressive models that adapts to the frequency sensitivity of each sample. 
we observe that there are existing step redundancy and unconditional branch redundancy in VAR. Motivated by this, we propose \textbf{SkipVAR}, a sample-adaptive framework that adaptively selects a frequency-aware acceleration strategy for each instance. Specifically, we first analyze the step redundancy during the generation and propose an automatic step-skip strategy to improve the efficiency. For unconditional branch redundancy, we discover that the information gap between the conditional and unconditional branch is small. Based on this observation, we introduce the unconditional branch replace, skipping the unconditional branch to decrease the computational cost. Extensive results demonstrate the effectiveness of frequency-aware training-free adaptive acceleration for efficient generation.


\newpage
\bibliographystyle{plain}
\bibliography{main}

\newpage
\appendix

\section{Scalability of Acceleration Strategies.}

Although our current acceleration strategy uses skip or unconditional branch replacement, more acceleration schemes can be integrated in principle. By leveraging the decision model to distinguish between high-sensitivity and low-sensitivity images, we can assign different acceleration levels accordingly—applying more aggressive acceleration to low-sensitivity images, while adopting more conservative strategies for high-sensitivity ones. This flexibility enables a finer-grained, image-aware control over inference efficiency, potentially unlocking greater acceleration gains while preserving output quality.

\section{Limitations of SSIM-Based Evaluation.}
In our experiments, we train decision models using SSIM-based thresholds to guide acceleration strategies. This objective metric helps align accelerated outputs with those generated by the original model. However, we acknowledge a key limitation: a lower SSIM score does not necessarily indicate perceptible quality degradation to the human eye. For instance, in face close-ups where high-frequency features dominate, even minor acceleration can significantly lower SSIM while leaving visual perception largely unaffected.

This observation suggests that relying solely on SSIM may lead to overly conservative decisions for images deemed high-frequency sensitive. To address this, we propose incorporating perceptual metrics that better reflect human visual perception to further differentiate within these “sensitive” images. Specifically, by identifying samples where visual quality remains acceptable despite low SSIM, we can apply more aggressive acceleration strategies without noticeable degradation. This approach enables us to unlock higher speedups while preserving the perceptual integrity of the outputs.

\section{Additional Generation Results on the Infinity-8B Model}

In this section, we provide further quantitative and qualitative results of applying our SkipVAR method on the Infinity-8B text-to-image model. Table~\ref{tab:drawbench} summarizes the DrawBench benchmarking results when using decision models trained on the 2B model but applied to the 8B model. 
\begin{table}[ht]
  \centering
  {\scriptsize
  \setlength{\tabcolsep}{4pt}
  \renewcommand{\arraystretch}{1.1}
  \rowcolors{2}{gray!10}{white}
  \caption{Quantitative Results on DrawBench using SkipVAR decision models trained on 2B, applied to Infinity-8B.}
  \label{tab:drawbench}
  \begin{tabular}{lcccccc}
    \toprule
    \rowcolor{gray!25}
    \textbf{Method}                     & \textbf{Latency} & \textbf{Speedup} & \textbf{SSIM} & \textbf{LPIPS} & \textbf{SSIM-HF} & \textbf{LPIPS-HF} \\
    \midrule
    Infinity\_8B                        & 3.38\,s          & ---              & ---           & ---            & ---              & ---               \\
    +SkipVAR@0.88\_via\_2B              & 1.86\,s          & $1.81\times$            & 0.901         & 0.0406         & 0.3885           & 0.1765            \\
    +SkipVAR@0.86\_via\_2B              & 1.74\,s          & $1.94\times$            & 0.8795        & 0.0476         & 0.3472           & 0.1857            \\
    +SkipVAR@0.84\_via\_2B              & 1.64\,s          & $2.06\times$           & 0.8592        & 0.0549         & 0.3121           & 0.1955            \\
    \bottomrule
  \end{tabular}
  }
\end{table}

\bigskip
On the Paintings dataset, which contains images with abundant high-frequency details, SkipVAR yields an average acceleration of $1.51\times$. Example generations showcasing diverse artistic styles are included in the \texttt{Images/} directory.

\begin{table}[ht]
\centering
\caption{Comparison of Super-Resolution (SR) and direct skipping for late steps in VAR model}
\begin{tabular}{lcccc}
\toprule
\textbf{Method} & \textbf{SSIM} ↑ & \textbf{LPIPS} ↓ & \textbf{SSIM$-\text{high}$} ↑ & \textbf{LPIPS$-\text{high}$} ↓ \\
\midrule
\rowcolor{gray!10}
SR-12         & 0.8412 & 0.1405 & 0.2128 & 0.3055 \\
Skip-12       & \cellcolor{blue!5}0.8621 & \cellcolor{blue!5}0.0632 & \cellcolor{blue!5}0.2596 & \cellcolor{blue!5}0.2011 \\
\rowcolor{gray!10}
SR-11-12      & 0.8078 & 0.1576 & 0.1841 & 0.3189 \\
Skip-11-12    & \cellcolor{blue!5}0.8218 & \cellcolor{blue!5}0.0875 & \cellcolor{blue!5}0.2099 & \cellcolor{blue!5}0.2233 \\
\rowcolor{gray!10}
SR-10-12      & 0.7712 & 0.1797 & 0.1599 & 0.3347 \\
Skip-10-12    & \cellcolor{blue!5}0.7802 & \cellcolor{blue!5}0.1172 & \cellcolor{blue!5}0.1726 & \cellcolor{blue!5}0.2467 \\
\bottomrule
\end{tabular}
\label{tab:sr_results}
\end{table}
\begin{figure}[ht]
  \centering
  \includegraphics[width=0.98\linewidth]{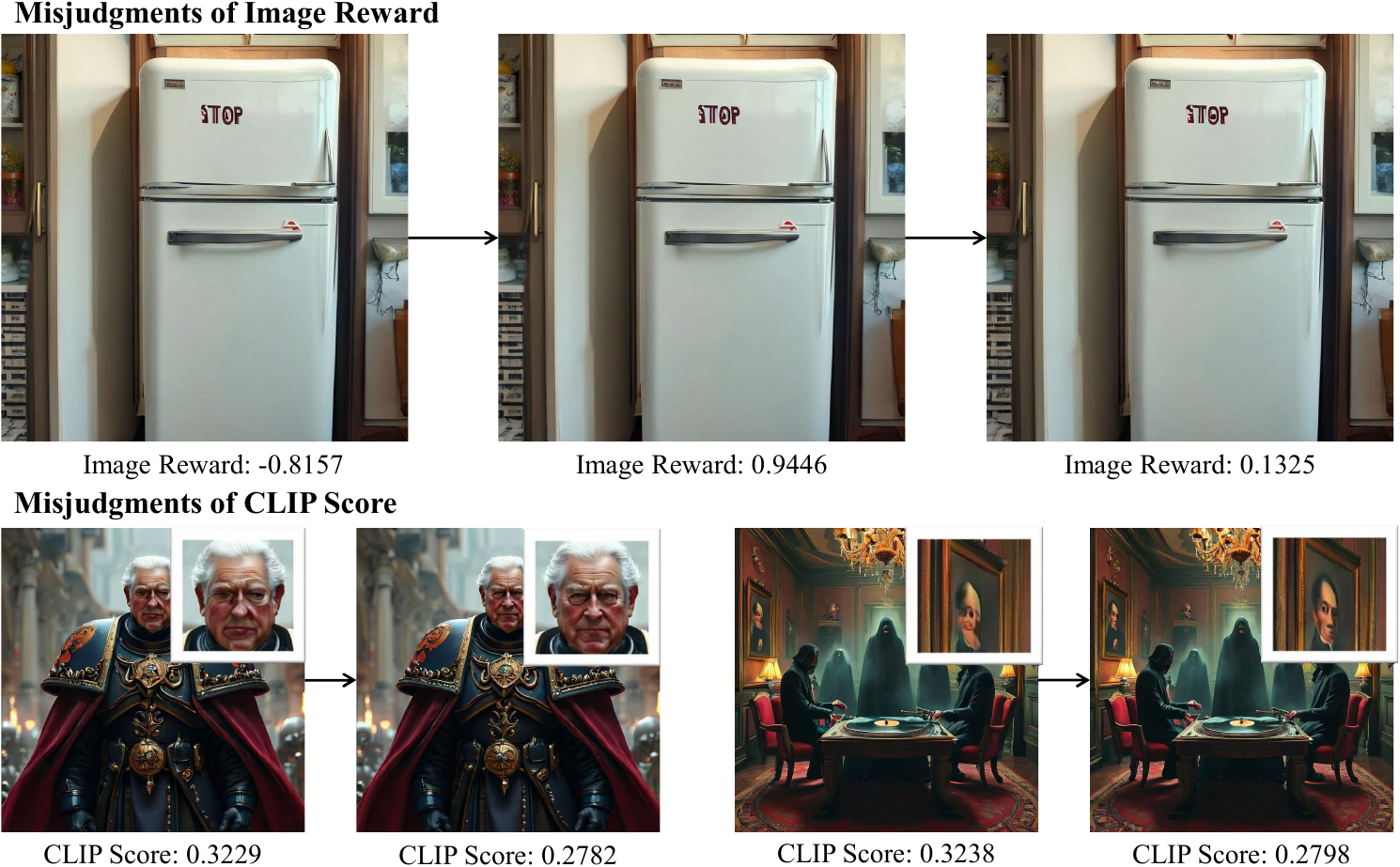}
  \caption{Misjudgments by ImageReward and CLIP Score on HF-Robust (top) and HF-Sensitive (bottom) images.}
  \label{fig:subjective_misjudgment}
\end{figure}
\section{Evaluating Step Replacement via Super-Resolution in VAR Models}
To investigate whether super-resolution (SR) models can compensate for the removal of later steps in the VAR model (specifically steps 10, 11, and 12), we conducted experiments comparing direct skipping and SR-based recovery methods. We adopt FreqFormer~\cite{wang2024freqformer} as the SR backbone in our evaluation. As shown in Table~\ref{tab:sr_results}, the use of super-resolution models (denoted with the "\_2x" suffix) results in consistently lower performance across multiple metrics such as SSIM, Cosine similarity, and LPIPS, when compared to their non-SR counterparts. Despite reducing inference time, the super-resolution models fail to effectively replicate the generation quality of the omitted steps. This suggests that directly replacing the later autoregressive steps with SR recovery cannot maintain the original model’s fidelity and thus does not bring meaningful improvements.

\section{Misjudgments of Common Subjective Metrics on Images}

We observe that widely used subjective evaluation metrics—namely ImageReward, CLIP Score, and GPT Score—often misjudge the importance of high‐frequency details. When evaluating images whose overall structure has been largely completed, these metrics fail to capture subtle but perceptually significant fine textures.

Specifically:
\begin{itemize}
  \item For \emph{HF-Robust} examples, human observers perceive negligible differences, yet ImageReward exhibit large fluctuations (see Figure~\ref{fig:subjective_misjudgment}).
  \item For \emph{HF-Sensitive} examples, the removal of high‐frequency details visibly degrades the image, but paradoxically the restoration of these details sometimes leads to a \emph{decrease} in the subjective scores.
  \item GPT Score exhibits significant instability: even when asked to distinguish between two images with clearly differing high‐frequency details—among a pool of 300 comparison images—GPT consistently fails to identify which image contains superior high‐frequency information.

\end{itemize}

These observations highlight the limitations of current subjective metrics in reflecting the contribution of fine visual details. Figure~\ref{fig:subjective_misjudgment} provides representative examples.
\end{document}